%% file: ms.tex
\documentclass{article}
\usepackage{ijcai20}

\usepackage{times}

\usepackage{soul}
\usepackage{url}
\usepackage[hidelinks]{hyperref}
\usepackage[utf8]{inputenc}
\usepackage[small]{caption}
\usepackage{graphicx}
\usepackage{booktabs}
\usepackage[ruled,linesnumbered]{algorithm2e}
\usepackage{mathtools}
\usepackage{amsmath,amssymb,amsfonts}
\usepackage{wasysym}
\usepackage{subfigure}
\DeclareMathOperator*{\argmin}{argmin}

\title{Multi-label Stream Classification with Self-Organizing Maps}

\author{
Ricardo Cerri$^1$\footnote{Contact Author}\and
Joel C. Junior$^1$\and
Elaine. R. F. Paiva$^2$\and
João M. P. Gama$^3$\\
\affiliations
$^1$Department of Computer Science - Federal University of S\~{a}o Carlos, S\~{a}o Carlos, Brazil\\
$^2$Federal University of Uberl\^{a}ndia, Uberl\^{a}ndia, Brazil\\
$^3$INESC TEC - University of Porto, Porto, Portugal\\
\emails
\{cerri, joel.costa\}@ufscar.br,
elaine@ufu.br,
jgama@fep.up.pt
}

\begin{document}
\DeclarePairedDelimiter{\ceil}{\lceil}{\rceil}
\SetKwInOut{Input}{Input}
\SetKwInOut{Output}{Output}

\maketitle

\begin{abstract}
Several learning algorithms have been proposed for offline multi-label classification. However, applications in areas such as traffic monitoring, social networks, and sensors produce data continuously, the so called data streams, posing challenges to batch multi-label learning. With the lack of stationarity in the distribution of data streams, new algorithms are needed to online adapt to such changes (concept drift). Also, in realistic applications, changes occur in scenarios of infinitely delayed labels, where the true classes of the arrival instances are never available. We propose an online unsupervised incremental method based on self-organizing maps for multi-label stream classification with infinitely delayed labels. In the classification phase, we use a k-nearest neighbors strategy to compute the winning neurons in the maps, adapting to concept drift by online adjusting neuron weight vectors and dataset label cardinality. We predict labels for each instance using the Bayes rule and the outputs of each neuron, adapting the probabilities and conditional probabilities of the classes in the stream. Experiments using synthetic and real datasets show that our method is highly competitive with several ones from the literature, in both stationary and concept drift scenarios. 
\end{abstract}

\input{introduction}

\input{related}

\input{method}

\input{methodology}

\input{experiments}

\input{conclusions}

%% The file named.bst is a bibliography style file for BibTeX 0.99c
\bibliographystyle{named}
\bibliography{biblio}

\end{document}

%% file: introduction.tex
\section{Introduction}
Multi-label Classification (MLC) is a machine learning task which associates multiple labels to an instance \cite{Tsoumakas2010}. This is a reality in many real-world applications such as bioinformatics, images, documents, movies, and music classification. 

Several works have addressed MLC in batch scenarios \cite{read2011classifier,nam2014large,Pliakos2018,Cerri2019}. They usually assume static probability distribution of data, and training instances being sufficiently representative of the problem. The decision model is built once and does not~evolve. 

Recent works in MLC bring a different scenario, where data flows continuously, in high speed, and with non-stationary distribution. This is known as data streams (DS)~\cite{gama2007learning}, bringing new challenges to MLC. Among them is concept drift, where learned concepts evolve over time, requiring constant model updating. Also, given the high velocity and volume of data, storing and scanning it several times is impractical. Many works have been developed to address such issues~\cite{read2010efficient,read2012scalable,song2014new,trajdos2015multi}%,faria2016MINAS}.

The first works in MLC for DS have addressed concept drift proposing techniques to update the model as new data arrives using supervised learning~\cite{read2012scalable,shi2014efficient,osojnik2017multi}. However, they assume that true labels of instances are immediately available after classification, which is an unrealistic assumption in several scenarios. 

Few works have addressed infinitely delayed labels to MLC for DS \cite{wang2012mining,zhu2018multi,CostaJunior2019}. They usually use k-means clustering to detect the emergence of new classes, updating the models in an unsupervised fashion, or use other strategies such as active learning, which do consider that some labels will be available some time. Also, these works are more focused in identifying the appearance of novel classes than in concept drift.

%having two phases, named offline and online. In the offline phase, labeled instances induce an initial decision model. In the online phase, this model classifies new unlabeled instances, which are used to update the model in an unsupervised fashion. Besides promising, the main  disadvantage is the previously definition of the number of clusters to represent a class. 

%Also, they consider the emergence of new classes, a phenomenon known as concept evolution, in a supervised fashion. When a sufficient number of labeled instances from new classes arrived, they are used to train/retrain the model. When a sufficient number of labeled instances from new classes arrived, they are used to train/retrain the model. 

%To consider realistic scenarios, recent works address the problem as a novelty detection task~\cite{faria2016novelty}. They usually use clustering techniques, having two phases, named offline and online. In the offline phase, labeled instances induce an initial decision model. In the online phase, this model classifies new unlabeled instances, which are used to update the model in an unsupervised fashion. Besides promising, there are disadvantages, such as the previously definition of the number of clusters, and the assumption of hyperspherical data. 

%definir proposta
This work proposes a different strategy to avoid the previously mentioned drawbacks of the existing methods. Instead of using k-means, we rely on self-organizing maps (SOMs). The neighborhood characteristic of the SOMs better explore the search space, forcing neurons to move according to each other, creating a topological ordering. As a result, the spacial location of a neuron corresponds to a particular domain or feature on the input instances. With this, we don't need to worry about the number of clusters, since the set of synaptic weights provide a good approximation of the input space~\cite{Haykin2009}.

%This work aims to fill a gap in the literature, proposing an unsupervised method for MLC for DS for infinitely delayed labels scenarios. Our method is based on self-organizing maps (SOM), 

Our proposal detects concept drift by adjusting the weight vectors of the neurons which classify arrival instances. We also decide the number of predicted labels for an instance based on an adaptive label cardinality, a Bayes rule that considers the outputs of each neuron, and online adapting probabilities and conditional probabilities of the classes in the~stream. The method is totally unsupervised during the online arrival of instances.%, our proposal is probably the first one (or is among the first ones) to consider truly infinitely delayed labels scenarios.

% organização do trabalho
This paper is organized as follows. Section~\ref{sec:relWork} discusses the main related works. Section~\ref{sec:method} presents our proposal, and Section~\ref{sec:methodology} brings the experimental methodology. The results are discussed in Section~\ref{sec:exp}. Finally, Section~\ref{sec:conclusion} presents our conclusions and future research directions.

%Most of these works use a model composed of a set of micro-clusters, a summary data structure, which allows evolving the model without the true labels.

%% file: related.tex
\section{Related Work}
\label{sec:relWork}

\cite{osojnik2017multi} adapted a multi-target regression method, but it poorly adapts to concept drifts and has a high computational complexity. \cite{sousa2018multi} proposed the same strategy with two problem transformation methods, ML-AMR and ML-RR, but also with high computation complexity. To deal with computational complexity and high memory consumption, \cite{ahmadi2018label} proposed a label compression method combining dependent labels into single pseudo labels. A classifier is then trained for each one of~them.

%\cite{Nguyen2019a} proposed a Bayesian-based strategy, trying to consider label correlations between pairs of labels, and also the relationships between labels and features. The Hoeffding inequality and the label cardinality are used to learn the number of labels to be predicted for each instance. The updating process, however, considers that the ground true labels are available with the arriving instances.

\cite{Nguyen2019b} proposed an incremental weighted clustering with a decay mechanism to detect changes in data, decreasing weights associated to each instance with time, focusing more on new arrived instances. For the classification, only clusters with weights greater than a threshold are used to assign labels to instances. The method also uses the Hoeffding inequality and the label cardinality to decide the number of labels to be predicted for an instance. However, the clusters are updated considering that the ground true labels arrive with the instances in the stream. 

To our knowledge, \cite{wang2012mining} was the first work to deal with delayed labels. It is a label-based ensemble with Active Learning to select the most representative instances to continually refine classes boundaries. The authors argue that using an ensemble, and updating the classifiers individually, they preserve information of classes which do not change when concept drift is detected. However, label dependencies are not considered.

\cite{zhu2018multi} proposed an anomaly detection method for concept evolution and infinitely delayed labels. It has three processes: \textit{i)} classification, using pairwise label ranking, binary linear classifiers, and a function to minimize the pairwise label ranking loss; \textit{ii)} detection, using Isolation Forest together with a clustering procedure in order to detect instances which may represent the emergence of new classes; and \textit{iii)} updating, building a classifier for each new class according to an optimization function. Although able to detect new classes, the method has difficulties with concept drifts, since changes in the streams are considered~anomalies. %This makes difficult for the update phase to distinguish between concept drift and concept evolution.

\cite{CostaJunior2019} proposed MINAS-BR, a clustering-based method using k-means for novelty detection. Offline labeled instances induce an initial decision model. This model classifies new online unlabeled instances, which are used to update the model in an unsupervised fashion. The method also considers that instances that are outside the radius of the existing clusters represent novelty classes, and new models must be constructed for them. Although promising, focusing on novelty detection can generate many false positives, harming the performance for concept drift detection.

From all methods reviewed here, only \cite{wang2012mining}, \cite{zhu2018multi} and \cite{CostaJunior2019} consider infinitely delayed labels. Wang et. al., however, use Active Learning, and thus consider that, at some point, labeled instances will be available. Zhu et. al. show promising results for anomaly detection, but fail to detect concept~drift. Costa Júnior et. al., although focusing on novelty detection, is also proposed for concept drift. Thus, we included MINAS-BR in our experiments, only considering its concept drift detection~strategy.

%% file: method.tex
\section{Our Proposal}
\label{sec:method}

%In this section we present our method for online learning detection of concept drift in multi-label stream classification. %Our proposal is based on self-organizing maps (SOM) and is totally unsupervised during the online arrival of instances, which situates it as probably the first one (or among the first ones) to consider truly infinitely delayed labels scenarios.

%In this section we present our method for online learning and detection of concept drift in multi-label stream classification using self-organizing maps (SOM). 
Our proposal is divided in two phases: \textit{i)} offline, using a labeled dataset to train models, and \textit{ii)} online, classifying arrival instances in a completely unsupervised~fashion.%, adapting the models to concept drift. For the classification, we use a k-nearest neighbors strategy to obtain winning neurons in the SOM maps. %It is important to recall that our method does not detect the emergence of new classes (concept evolution). In real scenarios, potentially any number of classes can emerge in a stream, and instances can belong to an arbitrary number of classes (existing and new) simultaneously. This is an extremely challenging problem requiring some assumptions to be made in order to treat the problems. Thus, given that concept drift in multi-label problems is already a very challenging task, concept evolution will be left for future~works.

% ===========================
% Versão sem Algoritmo 1
% ===========================

In our offline phase, $n$ SOM maps with $d \times d$ neurons are trained to represent each of the $n$ known classes. Each training instance is formed by a tuple $({\bf x}_i,Y_i)$, with ${\bf x}_i$ representing the feature vector of instance $i$, and $Y_i$ its corresponding set of classes. After calculating the training set label cardinality, we compute two $n \times n$ matrices $P$ and $T$. $T$ has the total number of instances classified in a class $y_j$ and in a pair of classes $(y_j,y_n)$. Matrix $T$ is used to compute $P$, which has the relationships between classes. $P$ stores class probabilities $p(y_j)$ and class conditional probabilities $p(y_j|y_n)$ for each one of the $n$ known classes. Positions $T[j,n]$ and $P[j,n]$ have, respectively, the number of instances classified in the pair $(y_j,y_n)$, and the conditional probabilities $p(y_j|y_n)$. Similarly, $T[j,j]$ and $P[j,j]$ have, respectively, the number of instances classified in $y_j$, and the probability $p(y_j)$. These matrices are used in the online phase for classification of instances in the stream. To compute $p(y_j|y_n)$, we have the conditional distribution between $y_j$ and $y_n$ based on the Bayes theorem:
\begin{equation}
    p(y_j|y_n) = \frac{p(y_j,y_n)}{p(y_n)} = \frac{f(y_j,y_n)}{f(y_n)}
    \label{eq:condProb}
\end{equation}

In Equation~\ref{eq:condProb}, $f(y_n)$ and $f(y_j,y_n)$ are obtained from the labeled dataset (matrix $T$), with $f(y_n)$ the number of instances classified in class $y_n$, and $f(y_j,y_n)$ the number of instances classified in both classes $y_j$ and $y_n$.

The next step constructs $n$ subsets $X_{y_j}$, each one with the instances classified in class $y_j$. We then build a SOM map for each of these subsets applying the well-known batch implementation of the Kohonen maps~\cite{Kohonen2013}. It is more recommended for practical applications, since it does not require a learning rate, and convergences faster and safer than the stepwise recursive version~\cite{Kohonen2013}.

The batch algorithm first compares each of the ${\bf x}_i$ vectors to all $d \times d$ neurons of the map, which had their weight vectors ${\bf m}$ randomly initialized. Then, a copy of ${\bf x}_i$ is stored into a sub-list associated with its best matching neuron $n_b$ according to the Euclidean distance:
\begin{equation}
    n_b = \underset{b}\argmin{||{\bf x}_i - {\bf m}_b||}
    \label{eq:bestUnit}
\end{equation}

Given $N_b$ as the neighborhood set of a neuron $n_b$, we compute a new vector ${\bf m}_b$ as the mean of all ${\bf x}_i$ that have been copied into the union of all sub-lists in $N_b$. This is performed for every neuron of the SOM grid. Old values of ${\bf m}_b$ are replaced by their respective means. This has the advantage of allowing the concurrent computation of the means and updating over all neurons. This cycle is repeated, cleaning the sub-lists of all neurons and redistributing the input vectors to their best matching neurons. Training stops when no changes are detected in the weight vectors in continued iterations. To avoid empty neurons, or neurons with very few mapped instances, we discard the ones with less than four mapped~instances.

The next step associates an average output and a threshold value to each neuron of the SOM maps. The average output is obtained my mapping $X_{y_j}$ to $map_{y_j}$. For each neuron $n_b$, we get the $X_b$ instances mapped to it, and then calculate the average of the discriminant functions $averOut_b$ over these instances:
\begin{equation}
    averOut_b = \sum_{i \in X_b}exp(-||{\bf x}_i - {\bf m}_b||) / |X_b|
    \label{eq:discriminantFunction}
\end{equation}

%In Equation~\ref{eq:discriminantFunction}, Dist is the Euclidean distance between the codebook (weight vector) ${\bf m}_b$ associated to neuron $n_b$, and an instance ${\bf x}_i$ belonging to the set $X_b$ of instances mapped to $n_b$. 
Having the average output of a neuron $n_b$, we compute its threshold value, which is used in the online phase to decide if a new instance is classified in the class associated to the map containing $n_b$. For this, we consider that an instance mapped to $n_b$ was already classified in all the other classes, except class $y_j$ associated to $map_{y_j}$. This is calculated using the Bayes rule:
\begin{equation}
    p(y_j|Y,X_b) = p(y_j) \times \prod_{y_k \in Y}p(y_k|y_j) \times p(X_b|y_j)
    \label{eq:condProbaThreshold}
\end{equation}

As already seen, we obtain $p(y_j)$ and $p(y_k|y_j)$ from data. Since $p(X_b|y_j)$ is the probability of observing $X_b$ given $y_j$, we have $p(X_b|y_j) =averOut_b$ (Equation~\ref{eq:discriminantFunction}). We thus avoid to manually set a threshold to decide when a neuron classifies an instance. If $p(y_k|y_j) = 0$ in matrix $P$, we do not consider this value in the calculation, otherwise we would have $p(y_j|Y,X_b) = 0$.

The online (classification) phase is detailed in Algorithm~\ref{alg:online}. Given an incoming unlabeled instance ${\bf x}_i$, we map it to each $map_{y_j}$. For each map, we retrieve a sorted list with the closest neurons to ${\bf x}_i$ (Algorithm~\ref{alg:online}, step~\ref{alg:online:map}). We also store the index of the closest neuron for each map, together with its corresponding discriminant function output (Algorithm~\ref{alg:online}, steps~\ref{alg:online:win1} to \ref{alg:online:win2}).

\begin{algorithm}[tb!]
    \scriptsize
    \DontPrintSemicolon
    \Input{Multi-label data stream ($DS$)\;
    \hspace{1.3cm}Label cardinality $z$\;
    \hspace{1.3cm}List with $n$ SOM maps ($MAP$)\;
    \hspace{1.3cm}Class probability matrix ($P$)\;
    \hspace{1.3cm}Class totals matrix ($T$)\;
    \hspace{1.3cm}Total number of instances ($N$)\;
    \hspace{1.3cm}Average neuron outputs ($ANO$)\;
    \hspace{1.3cm}Neurons thresholds ($NT$)\;}
    
    \Output{Updated $z$, $MAP$, $P$, $T$, $NT$, $N$, $ANO$\;
    }
    \Begin{
        %$t \leftarrow 0$ \tcp{timestamp}
        $kn \leftarrow minNumberNeuronsMAP(MAP)$\;
        \ForEach{instance ${\bf x}_i$ in $DS$}{
            $Y_i \leftarrow \emptyset$ \tcp{new prediction}
            %$t \leftarrow t + 1$\;
            $N \leftarrow N+1$\;
            $NrSort \leftarrow ListOfLists[[L_1],\dots,[L_n]]$\;
            $WinNr \leftarrow Array[1,\dots,n]$\;
            $outputWinNr \leftarrow Array[1,\dots,n]$\;
            \For{$j=1$ to $size(MAP)$}{
                $map_{y_j} \leftarrow MAP[j]$\;
                $NrSort[[j]] \leftarrow sortNeurons(map_{y_j},{\bf x}_i)$\;\label{alg:online:map}
                $WinNr[j] \leftarrow NrSort[[j]][1]$\;\label{alg:online:win1}
                ${\bf m} \leftarrow getWeightVector(WinNr[j])$\;
               % $distance \leftarrow eucDist({\bf m},{\bf x}_i)$\;
                $outputWinNr[j] \leftarrow exp(-||{\bf x}_i - {\bf m}||)$\;\label{alg:online:win2}
                
                %$D_{y_j} \leftarrow correspondMC(MC)$\;
                %$centroid \leftarrow correspondCentr(nr_{close})$\;
                %$distance \leftarrow eucDist(centroid,{\bf x}_i)$\;
            }
            $WinClasses \leftarrow getKNN(NrSort,kn)$\;\label{algo:online:knn}
            $c \leftarrow WinClasses[1]$\;\label{algo:online:winClass1}
            $Y_i \leftarrow Y_i \cup y_{c}$\;\label{algo:online:winClass2}
            \For{$k=2$ to $\ceil*{z}$}{\label{alg:online:classification1}
                $c \leftarrow WinClasses[k]$\;
                $p(y_c) \leftarrow P[c,c]$\;
                $p({\bf x}_i|y_c) \leftarrow outputWinNr[c]$\;
                $p(y_d|y_c) \leftarrow 1$\;
                \For{$l=1$ to $k-1$}{
                    $d \leftarrow WinClasses[l]$\;
                    \uIf{$y_d$ in $Y_i$}{
                        $p(y_d|y_c) \leftarrow p(y_d|y_c) \times P[d,c]$\;
                    }
                }
                $p(y_c|y_d,{\bf x}_i) \leftarrow p(y_c) \times p(y_d|y_c) \times p({\bf x}_i|y_c)$\;
                $tr \leftarrow NT[[c]][WinNr[c]]$\;
                \uIf{$p(y_c|y_d,{\bf x}_i) \geq tr$}{
                    $Y_i \leftarrow Y_i \cup y_c$\;
                }
            }\label{alg:online:classification2}
        $classifyInstance({\bf x}_i,Y_i)$\;    
        $MAP \leftarrow updateMAPs(MAP,Y_i)$\;
        $z \leftarrow updateLabelCardinality(z,Y_i,N)$\;
        $ANO \leftarrow updateAverNrOutputs(ANO,Y_i)$\;
        $T \leftarrow updateClassTotals(T,N,Y_i)$\;
        $P \leftarrow updateClassProbability(P,T,N,Y_i)$\;
        $NT \leftarrow updateThresholds(NT,P,ANO)$\;
        } 
        \KwRet{$(MAP, P, T, NT, ANO, N, z)$}\;
    }
    \caption{Online phase.}\label{alg:online}
\end{algorithm}
\setlength{\textfloatsep}{10pt}

\begin{figure}[t!]
    \centering
    \includegraphics[scale=1.1]{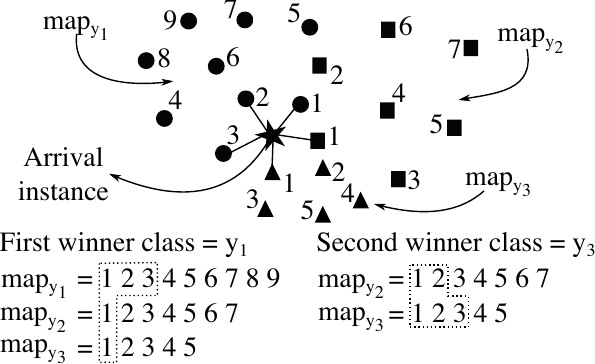}
    \caption{KNN procedure to select winning classes.}
    \label{fig:sortClasses}
\end{figure}

Given sorted lists $NrSort$ with the closest neurons to instance ${\bf x}_i$, we use a $k$-nearest neighbors strategy to retrieve a sorted list $WinClasses$ with the indexes of the winner classes of ${\bf x}_i$. Figure~\ref{fig:sortClasses} illustrates this (Algorithm~\ref{alg:online}, step~\ref{algo:online:knn}) for three maps with a maximum of nine neurons (grid dimension = 3), in a problem with three classes ($y_1, y_2, y_3$). In our proposal we always set $k$ as the number of neurons of the smallest map in $MAP$. If $k$ is even, we subtract 1 to guarantee an odd~number. In Figure~\ref{fig:sortClasses}, instance ${\bf x}_i$ is represented by a star ($\bigstar$). The other symbols represent the weight vectors of the neurons from $map_{y_1} (\newmoon)$, $map_{y_2} (\blacksquare)$, and $map_{y_3} (\blacktriangle)$. To get the winner class, we retrieve the $k=5$ nearest neurons from ${\bf x}_i$. We see that three of the closest neurons are from $map_{y_1}$, two from $map_{y_3}$, and one from $map_{y_2}$. From majority voting, class $y_1$ is the winner class. Neurons from $map_{y_1}$ are not considered anymore. It is easy to see now that from the five other closest neurons, three are from $map_{y_3}$ and two from $map_{y_2}$. The list $WinClasses$ then has the indexes 1, 3, 2 in this order. Now, ${\bf x}_i$ is classified in its closest class (Algorithm~\ref{alg:online}, steps~\ref{algo:online:winClass1} and \ref{algo:online:winClass2}), and the label cardinality $z$ is used to decide in which other classes to classify ${\bf x}_i$. We again use the Bayes rule and the class probabilities and conditional probabilities. Given a set $\hat{Y}_i$ with the classes in which ${\bf x}_i$ was already classified, the probability of classifying ${\bf x}_i$ in a new class $y_c$ is given by:
\begin{equation}
    p(y_c|\hat{Y}_i,{\bf x}_i) = p(y_c) \times \prod_{y_k \in \hat{Y}_i}p(y_k|y_c) \times p({\bf x}_i|y_c)
    \label{eq:condProbaClassify}
\end{equation}

We again obtain $p(y_c)$ and $p(y_k|y_c)$ from data. The probability $p({\bf x}_i|y_c)$ is given by $exp(-||{\bf x}_i - {\bf m}_b||)$, which is the output of the best matching neuron $n_b$ from $map_{y_c}$. If $p(y_c|\hat{Y}_i,{\bf x}_i)$ is greater or equal than the threshold associated to $n_b$ (Equation~\ref{eq:condProbaThreshold}), ${\bf x}_i$ is classified in $y_c$. This whole procedure is shown in Algorithm~\ref{alg:online}, steps~\ref{alg:online:classification1} to~\ref{alg:online:classification2}.

After classifying an $N$th instance, we update the maps of the $|Y_N|$ classes where ${\bf x}_N$ was classified. For each map, the weight vector of the best matching unit to ${\bf x}_N$ is updated with a fixed learning rate $\eta = 0.05$:
\begin{equation}
    {\bf m}_{b_{N}} = {\bf m}_{b_{N-1}} + \eta \times ({\bf x}_N - {\bf m}_{b_{N-1}})
    \label{eq:updateWeight}
\end{equation}

The label cardinality $z_{N}$ of the stream is also updated after classifying an $N$th instance (Equation~\ref{eq:updateLC}). Recall that $N$ is the total number of instances in the stream.
\begin{equation}
    z_N = \frac{1}{N}\sum_{i=1}^N |Y_i| = \frac{1}{N}((N-1) \times z_{N-1} + |Y_N|)
    \label{eq:updateLC}
\end{equation}

The average output $averOut_b$ of the best matching neuron ${\bf m}_b$ in each map corresponding to the classes in $Y_N$ is also updated:
\begin{equation}
    averOut_{b_N} = averOut_{b_{N-1}} + exp(-||{\bf x}_N - {\bf m}_{b_N}||)
    \label{eq:uptadeANO}
\end{equation}

We then update matrix $T$, and use it to update $P$ according to Equation~\ref{eq:condProb}. Finally we update the threshold values for each neuron using Equation~\ref{eq:condProbaThreshold}.

%% file: methodology.tex
\section{Methodology}
\label{sec:methodology}

Table~\ref{tab:data} presents our datasets, with number of numeric attributes ($A$), classes $(Y)$, and label cardinalities ($z$) for the initial labeled set. We generated four spherical ones using the MOA framework~\cite{bifet2010moa}. The classes are represented by possible overlapped clusters, and any overlap of clusters is a multi-label assignment. We also used the \cite{read2012scalable} proposal to generated one spherical dataset and three non-spherical~ones.

\begin{table}[htbp]    
  \scriptsize
	\centering
	\setlength{\tabcolsep}{7pt}
	\caption{Characteristics of the used datasets.}
		\begin{tabular}{l l l l l l}
		\toprule
			{Name} & $|DS|$ & $|A|$ & $|Y|$ & $z$ & $sd$\\
		\midrule	
		    Mult-Non-Spher-WF  & 100,000 & 21 & 7 & 2.37 & --\\
		    Mult-Non-Spher-RT & 99,586 & 30 & 8 & 2.54 & --\\
		    Mult-Non-Spher-HP  & 94,417 & 10 & 5 & 1.68 & --\\
		    Mult-Spher-RB & 99,911 & 80 & 22 & 2.24 & --\\
		    MOA-Spher-2C-2A & 96,907 & 2 & 2 & 1.06 & 1,000\\
		    MOA-Spher-5C-2A & 95,529 & 2 & 5 & 1.54 & 1,500\\
		    MOA-Spher-5C-3A & 94,667 & 3 & 5 &  1.37 & 1,500\\
		    MOA-Spher-3C-2A & 93,345 & 2 & 3 &  1.76 & 2,000\\
			Mediamill & 41,442 & 120 & 15 &  3.78 & --\\
			Nus-wide & 162,598 & 128 & 7 &  1.71 & --\\
			Scene & 1,642 & 294 & 4 &  1.07 & --\\
			Yeast & 2,364 & 103 & 9 &  4.15 & --\\
			%Reuters & 5,694 & 500 & 35 &  1.36 \\
		\bottomrule
		\end{tabular}
	    \label{tab:data}
\end{table}

The MOA datasets were generated with a radial basis function, where clusters are smoothly displaced after $sd$ instances in the stream. The dataset MOA-Spher-2C-2A has the additional characteristic that its clusters are simultaneously rotated around the same axis, moving close and away from each other. We used four generators with the multi-label generator: wave-form (WF), random tree (RT), radial basis function (RB), and hyper plane (HP). We varied their label relationships, which can influence label cardinalities. Label relationships are closely related to label skew (where a label or a set of labels is dominant in data). Thus, $p(y_k|y_j)$ is high if $p(y_k)$ is high, and low when $p(y_k)$ is low. We divided the stream $DS$ in four sub-streams. To insert concept drift, $10\%$ of the $p(y_k|y_j)$ values in the second and third sub-streams receive normally distributed random numbers with $\mu = p(y_k)$ and $\sigma = 1.0$. A value of $30\%$ is used in the fourth sub-stream. In all synthetic datasets, the initial 10\% of the stream is used for training ($D_{tr}$). A detailed description on how the pairwise relationships are generated is given by~\cite{read2012scalable}.

\begin{figure*}[htpb]
       \center
       \subfigure[refa][MOA-Spher-5C-2A]{\includegraphics[scale=0.5]{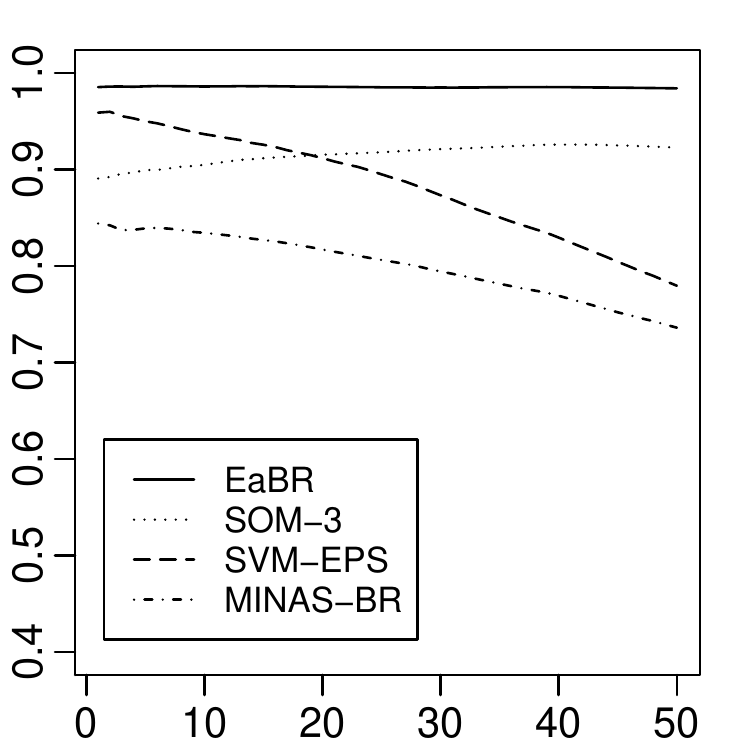}}
       \hspace{0.3em}
       \subfigure[refb][MOA-Spher-5C-3A]{\includegraphics[scale=0.5]{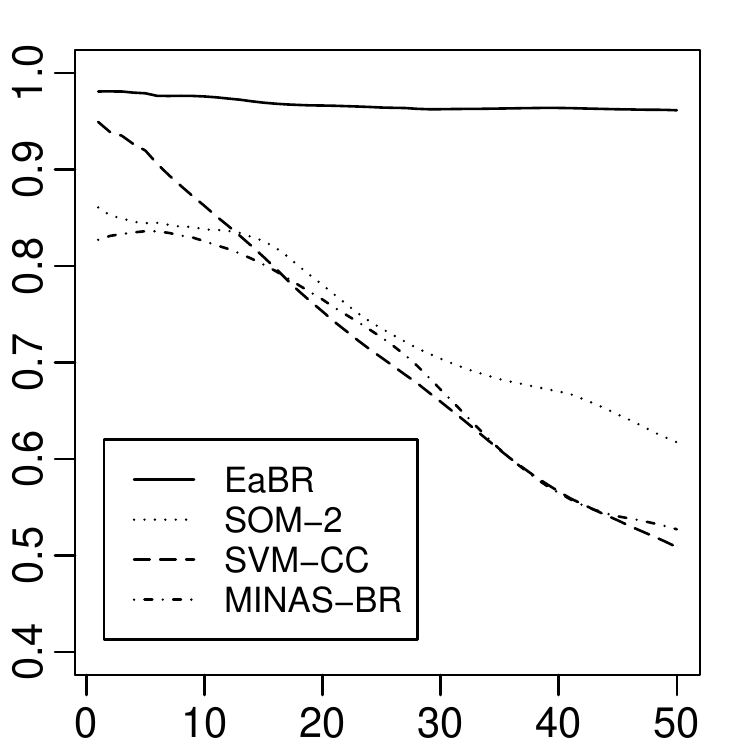}}
       \hspace{0.3em}
       \subfigure[refc][MOA-Spher-2C-2A]{\includegraphics[scale=0.5]{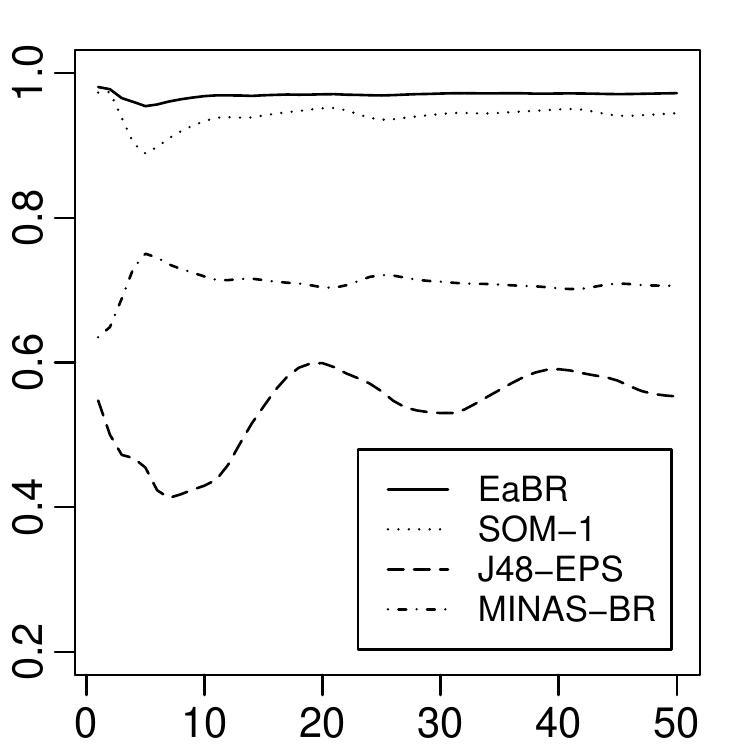}}
       \hspace{0.3em}
       \subfigure[refd][MOA-Spher-3C-2A]{\includegraphics[scale=0.5]{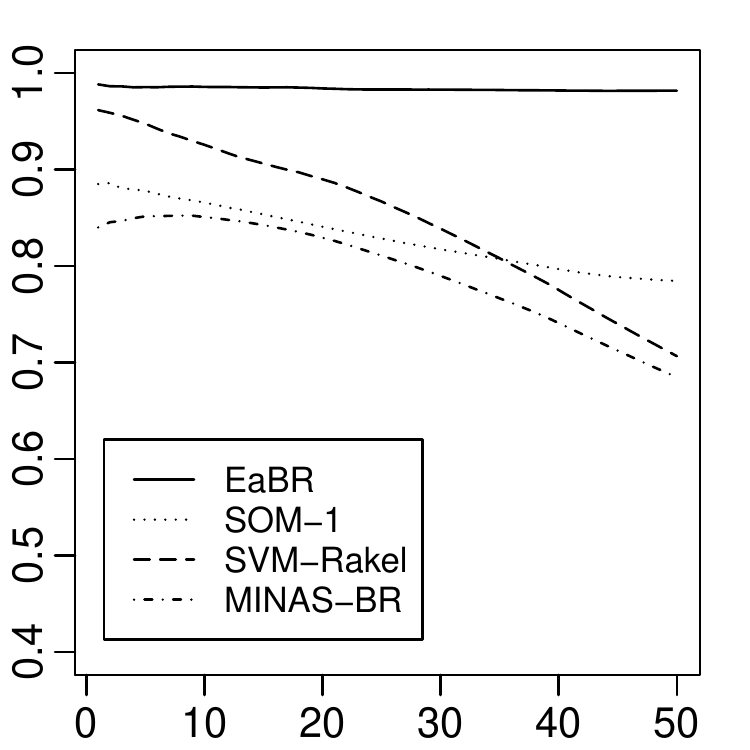}}
       %\caption{Results for the synthetic datasets generated with the MOA generator.}
       %\label{fig:MOAdata}
%\end{figure*}
%\begin{figure*}[tp]
%       \center
        \hspace{0.3em}
       \subfigure[ref1][Mult-Non-Spher-HP]{\includegraphics[scale=0.5]{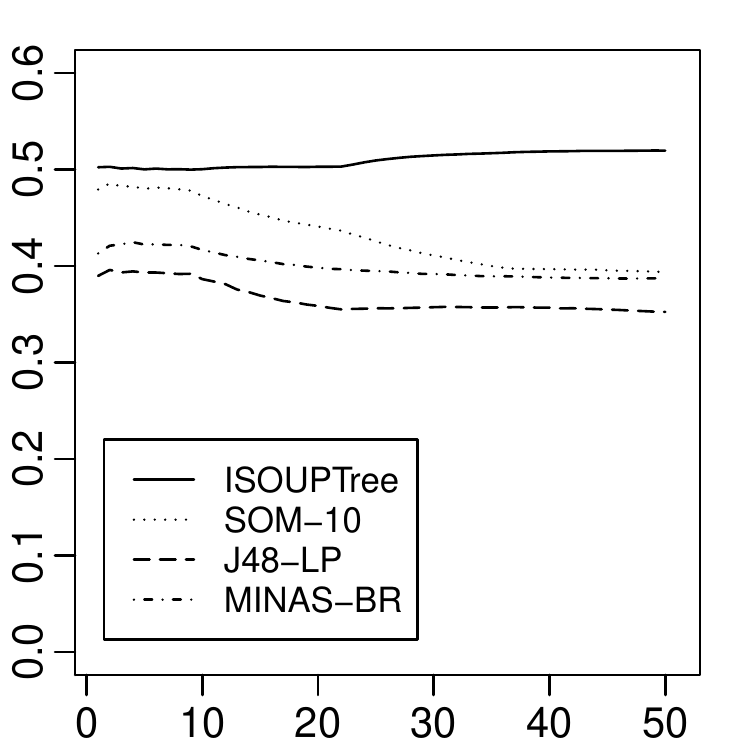}}
       \hspace{0.3em}
       \subfigure[ref2][Mult-Non-Spher-RT]{\includegraphics[scale=0.5]{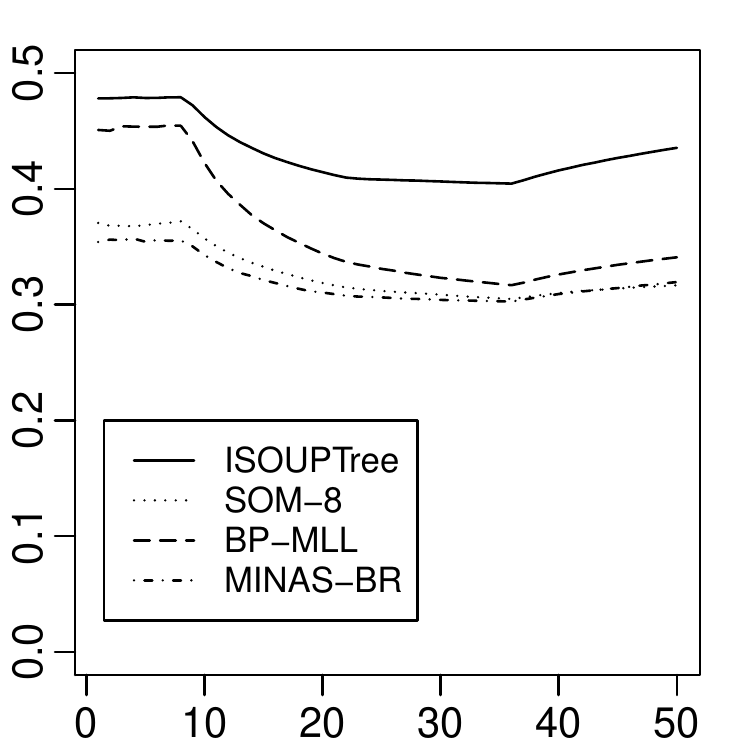}}
       \hspace{0.3em}
       \subfigure[ref2][Mult-Non-Spher-WF]{\includegraphics[scale=0.5]{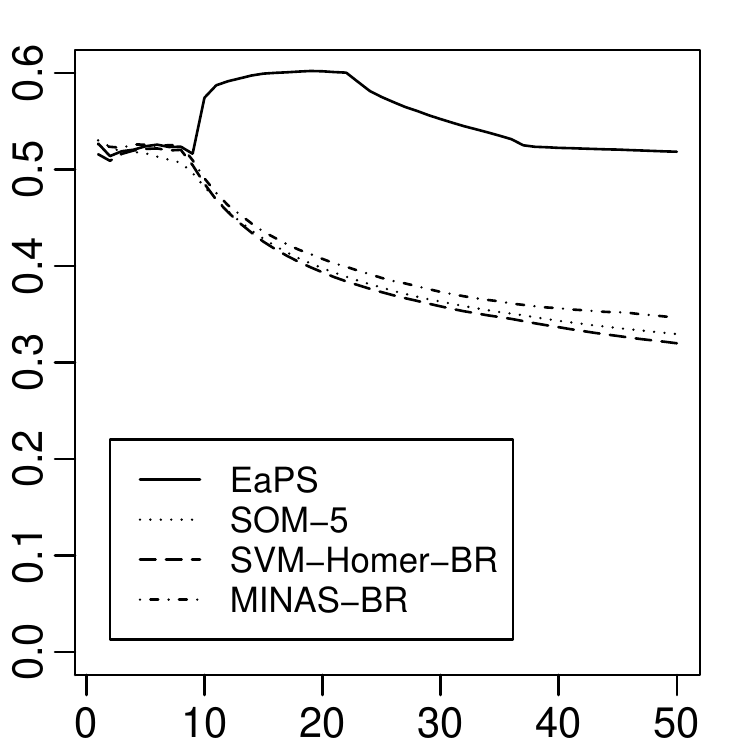}}
       \hspace{0.3em}
       \subfigure[ref2][Mult-Spher-RB]{\includegraphics[scale=0.5]{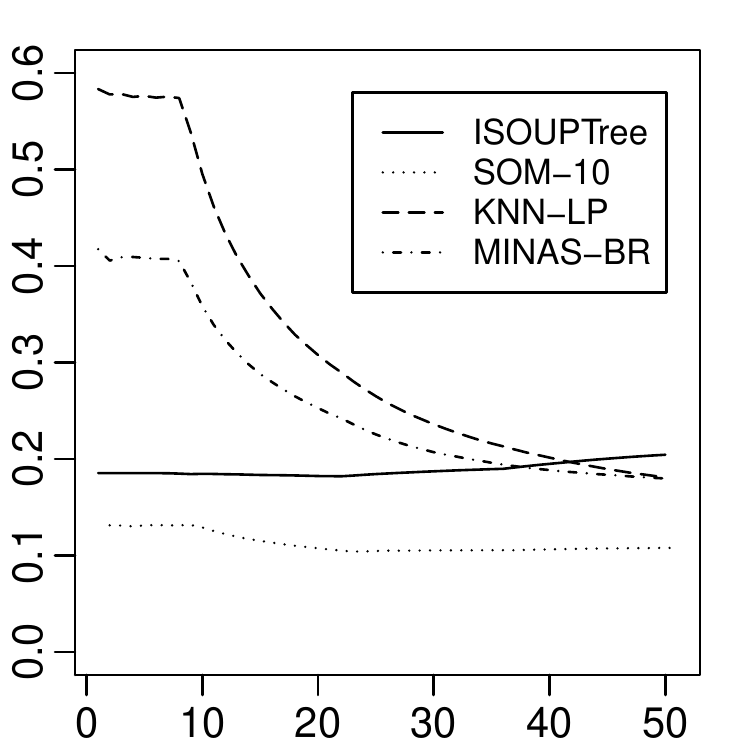}}
%       \caption{Results for the synthetic datasets generated with the Read et. al. generator.}
%       \label{fig:MultiData}
%\end{figure*}
%\begin{figure*}[tp]
%       \center
       \hspace{0.3em}
       \subfigure[ref1][Mediamill]{\includegraphics[scale=0.5]{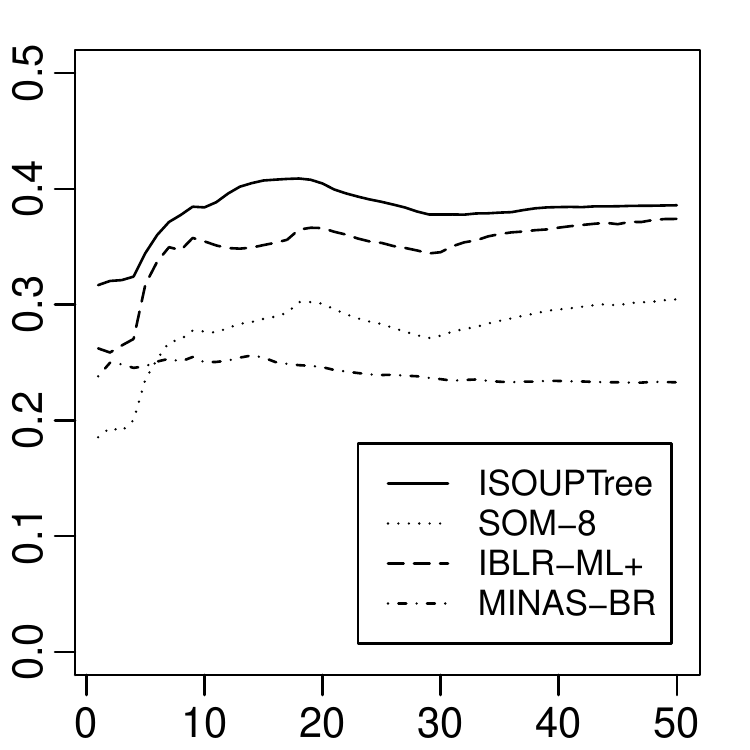}}
       \hspace{0.3em}
       \subfigure[ref2][Scene]{\includegraphics[scale=0.5]{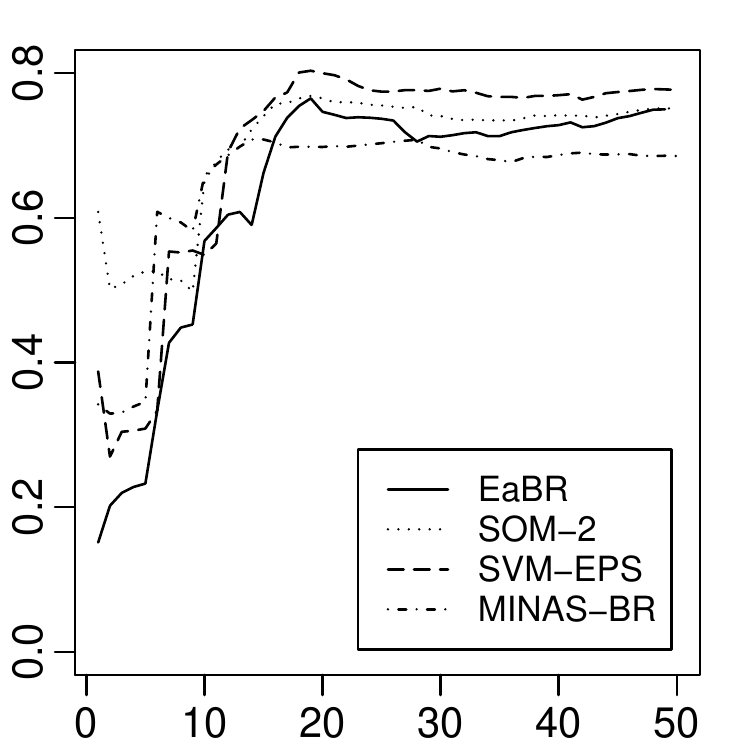}}
       \hspace{0.3em}
       \subfigure[ref2][Yeast]{\includegraphics[scale=0.5]{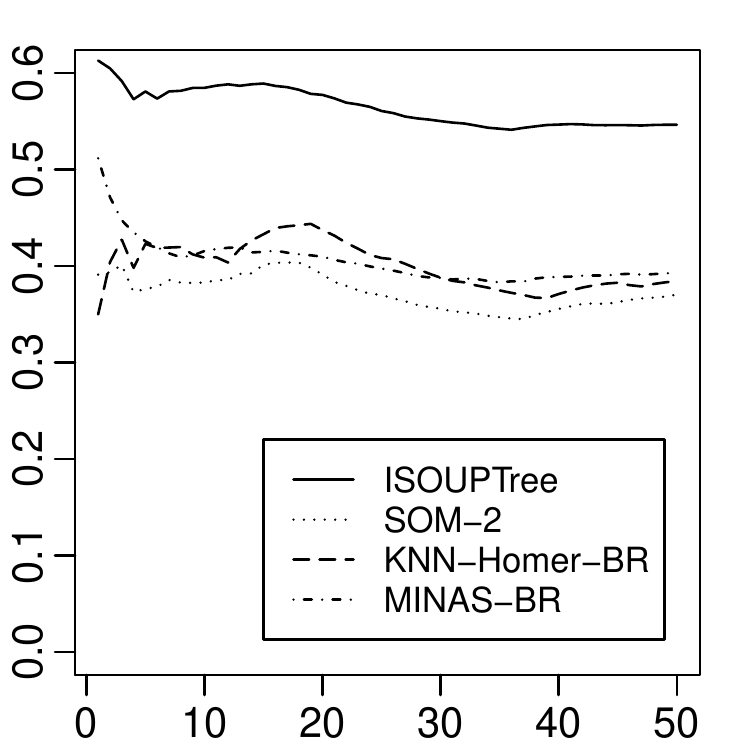}}
       \hspace{0.3em}
       \subfigure[ref2][Nus-wide]{\includegraphics[scale=0.5]{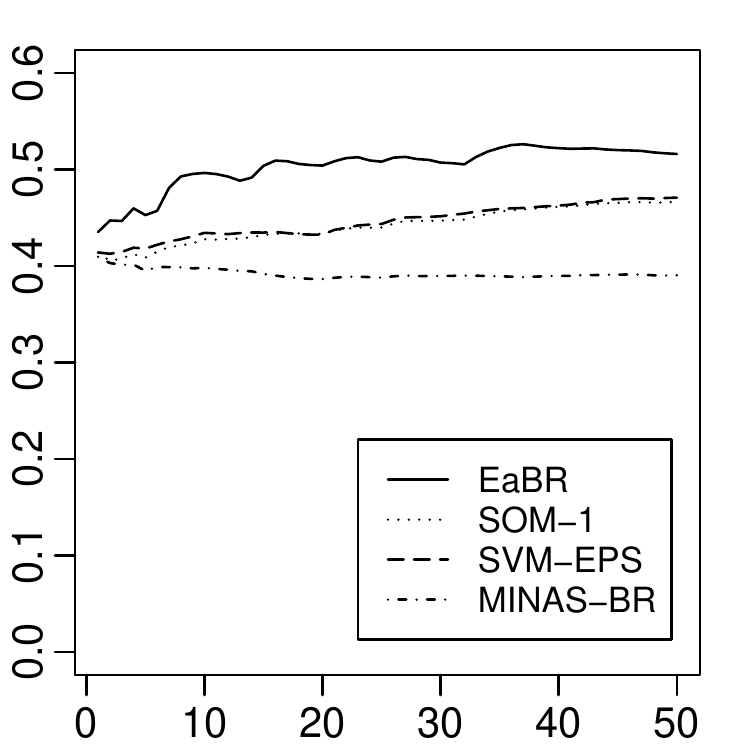}}
       \caption{Best results for all investigated datasets. Macro f-measure values over 50 evaluation windows.}
       \label{fig:results}
\end{figure*}

The four real datasets are from the Mulan website\footnote{http://mulan.sourceforge.net/datasets-mlc.html}. They are originally stationary, and were pre-processed to remove labels with less than 5\% of positive instances. The training set was constructed with 10\% of the data, trying to keep a same number of instances for each class.

We used 42 multi-label methods as baselines, with 31 being batch offline from Mulan~\cite{Mulan2011}, and 10 being online incremental from the MOA framework~\cite{bifet2010moa}. The Mulan methods are considered lower bounds, since they are trained with the offline dataset and are never updated. The MOA methods are considered upper bounds, since they are always incrementally updated using the true labels of the arrival instances. We also included MINAS-BR~\cite{CostaJunior2019}, up to now the only multi-label method in the literature which truly considers infinitely delayed labels.

We used problem transformations as lower bounds: Binary Relevance (BR), Label Powerset (LP), Randon k-Labelsets (Rakel), Classifier Chains (CC), Pruned Sets (PS), Ensemble of PS (EPS), Ensemble of CC (ECC), and Hierarchy of Multi-label Classifier (Homer), all with J48, SVM, and KNN as base classifiers. We also used algorithm adaptations: Multi-label KNN (ML-KNN), Multi-label Instance-Based Learning by Logistic Regression (IBLR-ML and IBLR-ML+), and Backpropagation for Multi-label Learning~(BP-MLL). BR, CC and PS with their ensembles were also used as upper bounds. We also used Multilabel Hoeffding Tree with PS (MLHT) and Incremental Structured Output Prediction Tree (ISOPTree), with their ensembles. They all use incremental Hoeffding Trees as base classifiers.

%% file: experiments.tex
\section{Experiments and Discussion}
\label{sec:exp}

\begin{figure*}[htbp]
     \centering
    \includegraphics[scale=0.76]{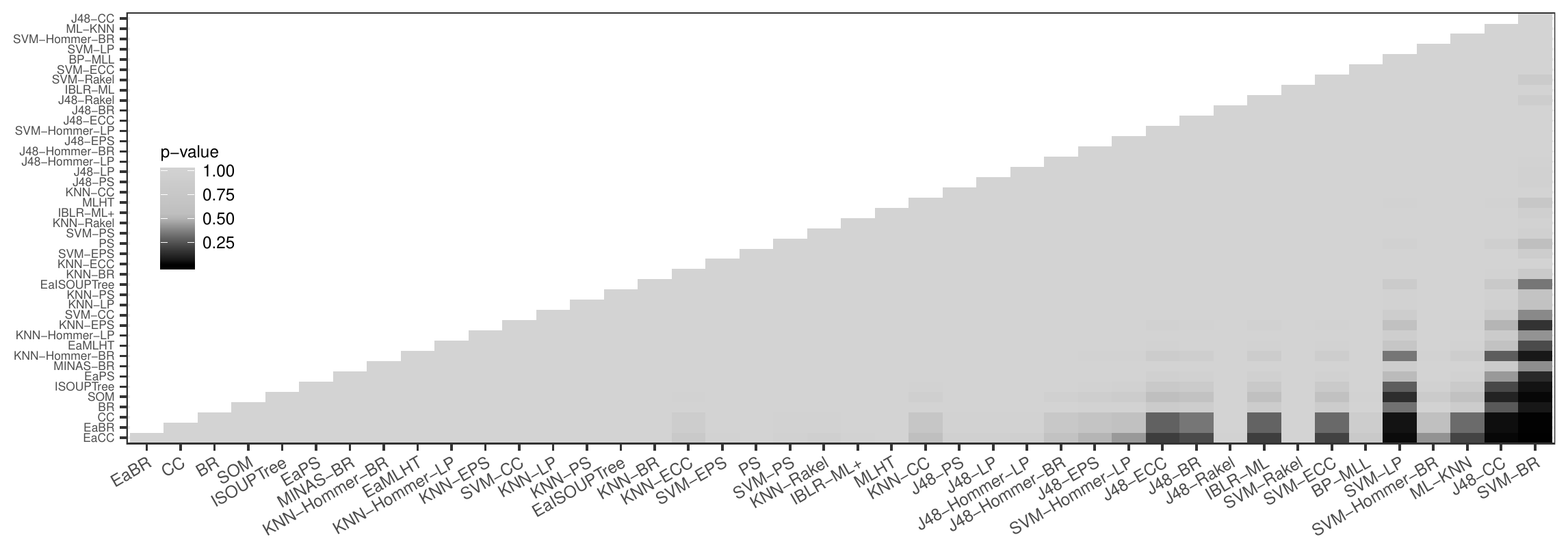}
    \caption{Methods' ranking and results of the Nemenyi statistical test. Both axes show all investigated methods.}
    \label{fig:heatmap}
\end{figure*}

Due space restrictions, Figure~\ref{fig:results} shows the best SOM, upper bound and lower bound results, and MINAS-BR. %All other results, datasets, and the proposal are freely available\footnote{Provided upon acceptance}.
We show multi-label macro f-measures (y-axix) across the entire stream over 50 evaluation windows (x-axis). The acronym Ea differentiates upper bound ensembles from the lower bound ones. SOM-$d$ refers to our proposal, with $d$ the dimension of the neurons grid (we used a hexagonal 2-$d$ grid in all experiments). We varied $d$ from 1 to 10 (1 to 100 neurons), executing each configuration 10 times in each dataset. We show the average results in each evaluation window, considering the SOM-$d$ with the high averages over the 50 evaluation windows. All other methods are deterministic, and were executed once. The exceptions were BP-MLL and MINAS-BR, which were executed 10~times. All 42 methods were executed with their default parameter~values.

The results for the MOA generated datasets (Figure~\ref{fig:results}(a-d) show that the performance of our proposal increased over the stream compared to the lower bounds and MINAS-BR, resulting in the best macro f-measures by the end of the stream. In MOA-Spher-5C-2A and MOA-Spher-2C-2A, we obtained very competitive results compared with the upper bounds. Since the MOA datasets are spherical, a small grid was enough to provide a good approximation of the feature space. In the datasets with two features, the 2-$d$ grid could obtain a more faithful representation of the input instances. This better maintained the topological ordering of the maps, {\it i.e.}, the spacial location of a neuron corresponded better to a particular feature from the input space. The clusters are well-behaved, and in some datasets only one neuron was enough to model a class. These characteristics combined with our proposed updating and kNN strategy resulted in a better adaptation to concept drift when compared to the lower bounds and~MINAS-BR.

In the non-spherical datasets generated with the Read et. al. generator (Figure~\ref{fig:results}(e-g)), our proposal performed similar to the lower bounds and MINAS-BR. We obtained a slightly better performance in Mult-Non-Spher-HP, being also very competitive in Mult-Non-Spher-RT and Mult-Non-Spher-WF. Differently from the spherical datasets, larger neuron grids were now necessary to better represent the input feature vectors. Although being spherical, the high number of features (80) combined with the high number of classes (22) in Mult-Spher-RB (Figure~\ref{fig:results}(h)) contributed to harm the performances of the SOMs. All methods, including the upper bounds, had generally worse performances in the Read et. al. generated datasets compared to the MOA generated ones. The former ones have more classes that are very overlapping, making the task much more difficult. All methods had difficulties in addapting to concept~drift.

Similar to the the results in the Read et. al. generated datasets, the results in the real datasets were generally worse than in the well-behaved MOA generated ones. Since there is no concept drift in these datasets, the batch algorithms obtained performances competitive to the upper bounds in the majority of the datasets. Our method was very competitive, being able to approach the upper bounds in two datasets.

Figure~\ref{fig:heatmap} presents a heat map pairwise comparing all 43 methods according to the post-hoc Nemenyi test~\cite{Demsar2006}, which was applied after the Friedman test returned a p-value = 6.181E-08. The figure also ranks the methods (x-axis left to right / y-axis bottom to top) according to their average macro f-measures over all datasets and evaluation windows. Our proposal was highly competitive to the baselines overall, being statistically equivalent. We obtained the fifth best performance, behind only the upper bounds EaCC, EaBR, CC and~BR. Very few statistically significant differences were detected, mainly between the top five methods of the ranking (including SOM) and the worst ranked ones, such as the BR transformation with SVM as base classifier. These differences are represented in Figure~\ref{fig:heatmap} by the black colored rectangles (p-values $<$ 0.05).

%% file: conclusions.tex
\section{Conclusions and Future Work}
\label{sec:conclusion}

In this work we proposed a novel method using self-organizing maps for multi-label stream classification in scenarios with infinitely delayed labels. Experiments on synthetically and real datasets showed that our proposal was highly competitive in different stationary and concept drift scenarios in comparison with batch lower bounds and incremental upper bounds. Our method takes advantage of the SOMs topology neighborhood behavior, forcing neurons to move according to each other in early stages of the training. This better exploitation of the search space, combined with our proposed updating and classification procedure, led to generally better results in comparison to MINAS-BR, up to now the only method which also considers infinitely delayed~labels.

Our method also has the advantage of having only two parameters, the learning rate for updating in the online phase, and the neuron grid dimension. However, we obtained very competitive results with a fixed learning rate. Our proposal can also be easily updated to eliminate the $d$ parameter by using dynamic versions of the SOM such as \cite{Alahakoon2000} and \cite{Dittenbach2000}.

As future works we will use dynamic self-organizing maps, and also extend our method to deal with concept evolution scenarios, where new classes can emerge over the stream. We also plan to investigate how to deal with structured streams, where classes are organized in topologies such as trees of graphs.